# The correlation between nativelike selection and prototypicality: a multilingual onomasiological case study using semantic embedding

Huasheng Zhang
Sichuan Normal University

huasheng.zhang@stu.sicnu.edu.cn

**Abstract:** In native speakers' lexical choices, a concept can be more readily expressed by one expression over another grammatical one, a phenomenon known as *nativelike selection* (NLS). In previous research, arbitrary chunks such as collocations have been considered crucial for this phenomenon. However, this study examines the possibility of analyzing the semantic motivation and deducibility behind some NLSs by exploring the correlation between NLS and prototypicality, specifically the onomasiological hypothesis of Grondelaers and Geeraerts (2003, Towards a pragmatic model of cognitive onomasiology. In Hubert Cuyckens, René Dirven & John R. Taylor (eds.), *Cognitive approaches to lexical semantics*, 67–92. Berlin: De Gruyter Mouton). They hypothesized that "[a] referent is more readily named by a lexical item if it is a salient member of the category denoted by that item". To provide a preliminary investigation of this important but rarely explored phenomenon, a series of innovative methods and procedures, including the use of semantic embedding and interlingual comparisons, is designed. Specifically, potential NLSs are efficiently discovered through an automatic exploratory analysis using topic modeling techniques, and then confirmed by manual inspection through frame semantics. Finally, to account for the NLS in question, cluster analysis and behavioral profile analysis are conducted to uncover a language-specific prototype for the Chinese verb *shang* 'harm', providing supporting evidence for the correlation between NLS and prototypicality.

Table of Contents

# 1 Introduction

Pawley and Syder (1983: 191) were the first linguists to call attention to the phenomenon of *nativelike selection* (NLS) as "the ability of the native speaker routinely to convey his meaning by an expression that is not only grammatical but also nativelike; what is puzzling about this is how he selects a sentence that is natural and idiomatic from among the range of grammatically correct paraphrases, many of which are non-nativelike or highly marked usages". For illustration, Pawley and Syder (1983) demonstrated that (1a) is more nativelike than (1b), (1c), etc.

(1)     a.   I want to marry you.

        b.   I wish to be wedded to you.

        c.   I want marriage with you.

        …

Over the years, *formulaic language* (e.g., collocations, chunks and multiword expressions) has been considered key to nativelike selection, including Pawley and Syder (1983) themselves (De Cock 1998; Erman 2009; Erman et al. 2016; Ortaçtepe 2013; Smiskova et al. 2012; Wray 2012). From previous research, the following two features can be summarized for NLS in the context of formulaic language. Due to this belief, it is advised by researchers and teachers in the second language classroom that these linguistic expressions should be taught and learned as an integral whole rather than being taken apart and analyzed in isolation for their underlying rules (Liu 2010; Woolward 2000):

i. *Arbitrariness*: This includes conventionality, idiomaticity and non-generativeness.

ii. *Chunkedness*: This includes multiwordness, (semi-)fixedness and holistic memorization.

However, there are opposing evidences and viewpoints challenging these two aspects:

i. Regarding *arbitrariness*: Alonso Ramos (2017), Bosque (2011), Liu (2010) and Walker (2008) questioned the assumed arbitrariness of some previously assumed collocations, highlighting their potential analyzability and deducibility (e.g., *\*powerful coffee* does not necessarily prove the arbitrariness of *strong coffee* when considering the nuanced semantic distinctions between *powerful* and *strong*). This deducibility is usually analyzed from a "predicate-driven" perspective rather than the traditional "from argument/base to predicate/collocate" direction (Almela-Sánchez 2019). Nesselhauf (2005: 32, 35) also noted the challenge in categorizing collocations, pointing out that some combinations can be interpreted as either arbitrary or not (e.g., *?commit a lie* may not affirm the arbitrariness of *commit a suicide* if the latter is viewed as a stand-alone sense of *commit*).

ii. Regarding *chunkedness*: Even though Foster (2009) and Zaytseva (2016) consider NLS as formulaic language, they also observed "nativelike selection of individual words". For instance, learners tend to use broad-meaning basic vocabularies (e.g., *understand*) in contrast to native speakers, who often prefer narrowly defined terms (e.g., *notice*, *realize*, *pay attention* and *care*). This indicates that NLS includes nuanced selection of individual words.

These findings suggest that NLS goes beyond merely formulaic language (i.e., arbitrary word combinations). This issue can be addressed through the onomasiological hypothesis proposed by Geeraerts et al. (1994) and, Grondelaers and Geeraerts (2003). They highlight prototypicality as the underlying reason why a lexical item is proactively selected for denoting a concept, rather than resolving to arbitrariness and chunkedness.

Research into how and why a concept is expressed by different linguistic items is relatively uncommon (Geeraerts 2015: 242). Recently, this onomasiological hypothesis is supported by Liu (2013) and Mehl (2018). New supporting evidence will be covered in this study.

The contributions of this study are:

i. A series of efficient, useful and innovative methods and procedures for exploring the correlation between nativelike selection and prototypicality is designed. These involve automatic explorations using semantic embedding, and manual analyses based on frame semantics and behavioral profile.

ii. This research is multilingual, offering insights from both intralingual and interlingual perspectives (Liu 2010). Also, It can better cope with the interference on word frequencies from extralinguistic real-world factors (Bosque 2011; Kjellmer 1991: 114).

iii. The research object is closer to people's everyday language life (i.e., customer reviews). This is important from a language learner perspective, and this perspective is crucial for NLS studies. The results demonstrate that NLS is vital for conveying basic concepts closely associated with people's real-world daily language life, and different languages prefer surprisingly different NLSs.

iv. The correlation addressed here is non-obvious and more implicit, enriching previous studies. Deeper semantic analysis is involved because the Chinese verb *shang* for the target NLS *shang shou* 'harm the hand' displays a language-specific prototype.

This study can be divided into 2 parts.

i. **NLS discovery** (Section 3): Multilingual customer reviews on a same product (i.e., dish soap) were collected from online marketplaces. Topic modeling leveraging semantic embedding (i.e., BERTopic) was used to discover potential NLSs automatically. Then, the target NLS was verified manually under the theory of frame semantics.

ii. **Prototype exploration** (Section 4): To support the correlation, a language-specific prototype of the target expression was revealed by conducting a manual behavioral profile analysis on the automatic clustering result of semantic embeddings.

Collectively, this case study sheds light on a new picture of NLS, a crucial language phenomenon for both learners and language research. By adopting the semantic explana-

**4 / 35**

tion of NLS-prototypicality correlation, it is easier to organize and depict language usages (Alonso Ramos 2017).

## 2 NLS from an onomasiological perspective and the hypothesis

The definition of NLS in this study is as follows:

Nativelike selection is the phenomenon that one expression is favored by native speakers over other grammatical and acceptable expressions that serve the equivalent function in discourse.

Before introducing the onomasiological perspective of NLS, I first clarify the formulaic language perspective of NLS. NLS, formulaic language and other terms in the field of formulaic language have been used and defined roughly interchangeably in research. This makes sense since formulaic language is a field that is notable for having as many as 60 alternate terms for it (Wray 2002: 8).

For example, "prefab" in Erman and Warren (2000: 31) is defined as "a combination of at least two words favored by native speakers in preference to an alternative combination which could have been equivalent had there been no conventionalization". A specially coined term is "conventionalized ways of saying things (CWOSTs)" which also includes exclusively "multi-word expressions" in Smiskova et al. (2012). At their core, these terms align with the definition of NLS in this paper, except for their stress on convention/arbitrariness and multiwordness/chunkedness. A radical opinion from this perspective is that "[a]ll lexical items are arbitrary - they are simply the consensus of what has been institutionalised" (Lewis 1997: 17).

However, as already discussed, arbitrariness and chunkedness do not apply to every NLS. Thus, NLS is addressed in this study from an onomasiological perspective. Grondelaers and Geeraerts (2003: 69, 70) introduce that "onomasiology takes its starting-point in a concept, and investigates by which different expressions the concept can be designated, or named" and it deals with questions "What are the relations among the alternative expressions?" and "What factors determine the choice for one or the other alternative?". The latter question is the central concern in this paper. In plain language, onomasiology concerns "Why does someone use the category x rather than the category y for talking about phenomenon z" (Geeraerts 2006: 27). Apparently, unlike the formulaic language perspective, the term onomasiology is neutral in meaning and does not presuppose arbitrariness and chunkedness.

The hypothesis under investigation in this paper is:

"A referent is more readily named by a lexical item if it is a salient member of the category denoted by that item" (Grondelaers and Geeraerts 2003: 74).

This delineates the empirical findings in Geeraerts et al. (1994). The term used in their studies is the correlation between prototypicality/semasiological salience and onomasiological cue validity/salience. Geeraerts et al. (1994) compiled a database of clothing photos paired with their corresponding referring words from fashion magazines, and discovered "In plain language: when you have to name something, you preferentially choose those items of which the thing to be named is a typical representative".

## 3 An exploratory analysis of NLS based on multilingual customer reviews

In what follows, before validating the correlation between NLS and prototypicality, the target NLS will be identified and confirmed from the corpus.

### 3.1 Automatic discovery of NLSs through BERTopic

#### 3.1.1 Method

Methods exploring NLS in previous research include corpus analysis (Geeraerts et al. 1994; Mehl 2018), forced-choice task (Liu 2013) and elicitation experiments aimed at either entire discourses (Foster 2009; Smiskova et al. 2012; Zaytseva 2016) or specific turns within discourses (Ortaçtepe 2013). All of them are carefully designed, to ensure that descriptions of the same topic(s) can be successfully collected and retrieved through manual scrutiny. The issue with these studies is that they require manual labor to design and inspect each topic in advance, before knowing whether possible NLSs can be successfully identified in the topics. Some even require designing the topics in advance and then collecting the data accordingly.

However, retrieving topics from a set of documents is a well-developed field in natural language processing, known as topic modeling, especially recently with the fast development of neural network-based language models under the emergence of attention mechanism and transformer architecture, which enable language models to have a nuanced understanding of the language context (Bahdanau et al. 2014; Devlin et al. 2018; Vaswani et al. 2017).

Therefore, different from previous research, an exploratory analysis aimed at the automatic discovery of potential NLSs is added before manual inspection, which is a combination of BERTopic (Grootendorst 2022) and OCTIS (Terragni et al. 2021) ).

- BERTopic leverages language models to generate a single vector (i.e., semantic embedding) for every single document. These documents are then grouped into different topics by clustering these vectors. Finally, topic words for each topic are calculated by c-TF-IDF. A vector can be imagined as a coordinate in a multidimensional space (e.g., [-0.12, 0.58, ...] with up to 1024 dimensions). The distance between any two vectors indicates the degree of their semantic similarity—the closer the vectors, the more similar they are. The clustering method uses a special combination of UMAP (McInnes et al. 2018) and HDBSCAN (Campello et al. 2013). Although this combination is controversial, it works well in practice (see the official documentation for the Python umap[1] package; Allaoui et al. 2020). For an introduction to semantic embedding, please refer to Lenci (2018).

- OCTIS is a framework that can be used to optimize the results of topic modeling methods by finding hyper-parameters that yield optimal evaluation metrics, using an efficient algorithm Bayesian optimization. The evaluation metric adopted here is NPMI (Bouma 2009) which is also used in Grootendorst (2022). NPMI measures how relatively frequently topic words in the same topic occur together in the whole corpus. The search space for optimized hyper-parameters of BERTopic include n_neighbors and n_components for UMAP, along with min_cluster_size and min_samples for HDBSCAN.

The language models used here are GTE-large-zh and GTE-large (Li et al. 2023) for Chinese and English, respectively, and sup-simcse-ja-large (Tsukagoshi et al. 2023) for Japanese. These models achieve leading performance in benchmarks of text embedding such as MTEB (Muennighoff et al. 2022). All the language models in this study were used out-of-the-box with their default configurations (e.g., choice of the output layer). This aims to align with the original training objectives (i.e., judging the semantic similarity of text pairs) of these models (Rogers et al. 2020).

### 3.1.2 Data collection and processing

The data was collected from customer reviews of a dish soap that sells internationally across China, the UK and Japan, to make the results comparable. The unique selling point of this product is its use of natural ingredients. Topics concerning this point can be expected in the collected data.

The customer reviews were collected from three online marketplaces: jd.com (China), amazon.co.uk (United Kingdom) and amazon.co.jp (Japan). To get more fine-grained topic representations, reviews were split into sentences, and each sentence was considered as an individual document because BERTopic assigns a single topic per document. There are difficulties for traditional topic modeling methods to handle short texts, but BERTopic is adept at short texts (de Groot et al. 2022; Egger and Yu, 2022). Finally, the removal of stopwords (e.g., *I*, *and* and *to*) and punctuations, and lemmatization were applied before calculating the topic words based on c-TF-IDF. Language parser tools used in this study were spaCy [2] for English, LTP (Che et al. 2021) for Chinese and GINZA[3] for Japanese.

The data structure is as follows: Chinese (65839 characters from 3487 sentences), English (23709 words from 2521 sentences) and Japanese (103116 characters from 3880 sentences).

---

[1] https://umap-learn.readthedocs.io/en/latest/clustering.html

[2] https://spacy.io/models/en_core_web_trf

[3] https://github.com/megagonlabs/ginza

### 3.1.3 Result and discussion

There are respectively 23, 17 and 23 topics extracted for Chinese, English and Japanese documents as the result of BERTopic after 150 Bayesian optimization iterations using OCTIS. Note that some documents are classified as outliers, and semantically similar documents may be separately grouped into adjacent clusters, resulting in separate topics. Due to limited space, for each language, only 4 topics are selected to be displayed here, and topic words are ordered by c-TF-IDF. Overall, these topic representations demonstrate semantic coherence within BERTopic topics and the feasibility of crosslinguistic comparison. These topics are manually labeled as `Overall opinion`, `Environmental concern`, `Scent` and `Skin feel`. The superscript number indicates the raw frequency of the topic or the word across the whole database. These frequencies should be compared with caution, as customers from different countries may focus on different topics.

**Chinese:**

`Overall opinion`$^{246}$: *bucuo*$^{507}$ 'not bad', *hao*$^{997}$ 'good', *haoping*$^{48}$ 'positive review, praise', *manyi*$^{84}$ 'satisfied', *ting*$^{171}$ 'quite', *hai*$^{332}$ 'fairly', *xing*$^{21}$ 'ok', *hen*$^{1074}$ 'very', *zan*$^{23}$ 'thumbs-up, awesome' and *feichang*$^{384}$ 'extremely'.

`Environmental concern` and `Overall opinion`$^{117}$: *huanbao*$^{240}$ 'environment-friendly', *fangxin*$^{121}$ 'not worry', *anquan*$^{101}$ 'safe', *chanpin*$^{253}$ 'product', *pinpai*$^{129}$ 'brand', *xinren*$^{61}$ 'trust', *hen*$^{1074}$ 'very', *haoyong*$^{372}$ 'easy-to-use; useful', *zhide*$^{107}$ 'worth' and *shou*$^{253}$ 'hand'.

`Scent`$^{256}$: *xiangwei*$^{167}$ 'fragrance', *weidao*$^{242}$ 'taste', *chanpin*$^{253}$ 'product', *meiyou*$^{216}$ 'there is not', *wu*$^{122}$ 'no', *xiang*$^{38}$ 'fragrance', *wuwei*$^{53}$ 'odorless', *haowen*$^{46}$ 'to smell good', *dandande*$^{32}$ 'faint (smell)' and *qiwei*$^{26}$ 'smell'.

`Skin feel`$^{135}$: *ganjing*$^{237}$ 'clean', *shou*$^{253}$ 'hand', *shang*$^{222}$ 'harm', *qingxi*$^{109}$ 'wash', *xi*$^{433}$ 'wash', *chongxi*$^{80}$ 'rinse', *bu*$^{601}$ 'not', *qingjie*$^{59}$ 'clean', *hao*$^{997}$ 'good' and *ganshou*$^{93}$ 'feeling'.

**English:**

`Overall opinion`$^{469}$: *product*$^{318}$, *good*$^{466}$, *thank*$^{28}$, *great*$^{279}$, *love*$^{93}$, *last*$^{101}$, *buy*$^{178}$, *time*$^{75}$, *job*$^{69}$ and *stuff*$^{32}$.

`Environmental concen`$^{138}$: *eco*$^{122}$, *environment*$^{64}$, *friendly*$^{97}$, *planet*$^{21}$, *kind*$^{49}$, *product*$^{318}$, *save*$^{41}$, *chemical*$^{30}$, *good*$^{466}$ and *animal*$^{9}$.

`Scent`$^{123}$: *smell*$^{199}$, *scent*$^{128}$, *lovely*$^{69}$, *nice*$^{80}$, *amazing*$^{36}$, *fragrance*$^{45}$, *like*$^{113}$, *long*$^{98}$, *great*$^{279}$ and *mild*$^{12}$.

`Skin feel`$^{121}$: *hand*$^{141}$, *skin*$^{159}$, *sensitive*$^{84}$, *gentle*$^{41}$, *dry*$^{45}$, *kind*$^{49}$, *sore*$^{12}$, *use*$^{246}$, *soft*$^{8}$ and *extremely*$^{9}$.

**Japanese:**

Overall opinion[661]: *tsukau*[703] 'use', *koonyuu*[129] 'purchase', *ii*[744] 'good', *shoohin*[160] 'product', *nedan*[53] 'price', *omou*[336] 'think', *tsuzukeru*[34] 'continue', *tai*[86] 'want', *ripi*[24] 'buy again' and *kau*[69] 'buy'.

Environmental concen[67]: *kankyoo*[135] 'environment', *eko*[82] 'eco (friendly)', *shizen*[52] 'nature', *yasashii*[82] 'gentle', *hairyo*[23] 'consideration', *yasashii*[346] 'gentle'[4], *yurai*[55] '(being made) from', *ii*[744] 'good', *kangaeru*[29] 'think' and *omou*[336] 'think'.

Scent[250]: *kaori*[452] 'fragrance', *nioi*[89] 'smell', *kooryoo*[82] '(aromatic) essence', *mu*[72] 'no', *ii*[744] 'good', *tsuyoi*[94] 'strong', *suki*[57] 'like', *kusai*[17] 'smelly', *yasashii*[346] 'gentle' and *nigate*[20] 'bad at'.

Skin feel[389]: *teare*[252] 'chapped hand', *hada*[207] 'skin', *areru*[133] 'chap', *te*[379] 'hand', *hadaare*[43] 'chapped skin', *binkan*[44] 'sensitive', *yasashii*[346] 'gentle', *hidoi*[55] 'severe', *tehada*[44] 'hand skin' and *shisshin*[35] 'eczema'.

A quick look at the results reveals that different languages seem to prefer different NLSs when expressing the same topic. The most prominent differences are found in Skin feel, where *shang*[222] 'harm', *sensitive*[84] and *teare*[252] 'chapped hand' are respectively favored in Chinese, English and Japanese (to praise the product's harmlessness to the skin). To express Overall opinion, *bucuo*[507] 'not bad', *haoping*[48] 'positive review, praise', *manyi*[84] 'satisfied', *hai*[332] 'fairly', *fangxin*[121] 'not worry', *anquan*[101] 'safe', *xinren*[61] 'trust' and *haoyong*[372] 'easy-to-use; useful' are common in Chinese but infrequent or even awkward in English and Japanese. Conversely, *love*[93] for Overall opinion, *lovely*[69] and *amazing*[36] for Scent are common in English but awkward in Chinese and Japanese. *Planet*[21] is used for Environmental concern but is awkward in Chinese and Japanese, whereas Japanese distinctively uses *yasashii*[428] 'gentle'. Other lexical gaps or language-specific words (some are not shown in this excerpt) include *zan*[23] 'thumbs-up, awesome', *nigate*[20] 'bad at (for expressing dislikeness)', *ripi/ripiito*[71] 'buy again' and *huigou*[176] 'buy again' (*repurchase* only occurred 2 times). Also, Chinese does not have a single word for *tsumekae*[99]/*refill*[43].

The reason these expressions are NLSs can be simply attributed to their special meanings rather than collocational chunks. This is because their usages are semantically transparent and do not have strong collocational dependency/fixed patterns or semantic cohesion (Alonso Ramos 2017) with specific words or topics. Some words are even hard to translate into another language.

Finally, the presence of collocations in the analysis is also acknowledged. When referring to bubbles, while English uses *lather*[26] or *foam*[27], Japanese and Chinese opt for *awadatsu*[267] 'bubble rises' and *fengfu*[33] 'abundant' respectively. When referring to removing grease, Chinese also uses *qu*[84] 'remove', but Japanese employs *ochiru/otosu*[399] 'drop'. In the framework of *lexical function* (Mel'čuk 1998), these can be analyzed as Magn(*paomo* 'bubble') = *fengfu* 'abundant', CausPredMinus(*you* 'grease') = *qu* 'remove', etc. Other collocations include *zhide xinlai*[44] 'worth trusting', *eco/environmentally friendly*[87], *kankyoo ni hairyo suru*[19] 'care the environment' and *shokubutsu yurai*[44] '(made) from plants' which have distinctively higher frequencies than their translations in other languages.

---

[4] The automatic Japanese parser distinguished two orthographic forms for *yasashii.*

The results appear efficient and successful as an automatic exploratory analysis to find potential NLSs. If done manually, it is hard and time-consuming to design, find and then count the recurring topics and the related words that are scattered across numerous documents. For example, the topic Skin feel can be referred to with various words (*skin*, *body*, *hand*, etc.) or even without mentioning body parts by simply using *soft*, *no allergies*, *no need to wear gloves*, etc.

### 3.2 Manual inspection based on frame semantics

Among the topics identified in Section 3.1.3, Skin feel (and its potential NLSs like Chinese *shang* 'harm') will be selected for detailed manual inspection. This will validate the automatic results, uncover more detailed information, and provide guidance on how to interpret the automatic results. Skin feel is chosen because the results show that distinctly different NLSs are used crosslinguistically in this topic, and the extracted topic words are semantically coherent and well-formed in at least English and Japanese.

### 3.2.1 Method

In this study, I introduce a new approach based on frame semantics for the manual inspection of NLSs. It provides an effective framework for comparing and categorizing various NLSs and their variants within a topic, and facilitates quantitative analysis. For example, Liu (2010) noted the difficulties of interlingual NLS comparison due to difficulties of comparing the meanings of crosslinguistic words, which can, however, be handled by frame semantics.

The concept of *frame*, initially introduced by Fillmore (1982), can serve as a universal "interlingua" for intralingual and interlingual descriptions and comparisons (Boas 2005, 2020; Hasegawa et al. 2011; Schmidt 2009). One main reason is that the perspective of frames is from semantics (Baker and Ruppenhofer 2002), abstracting away the morphosyntactic differences (Hasegawa et al. 2014). Therefore, interlingual applications of semantic frames include the uncovering of language-specific words (Schmidt 2009; VanNoy 2017), and the objective evaluation/identification of translation accuracy/divergence (Ellsworth et al. 2021; Hasegawa et al. 2014). Such divergence can be termed as "frame shift" (Yong et al. 2022).

In line with the framework of Ruppenhofer et al. (2016), this paper views Skin_feel as a "Super frame", which has five "Perspective_on" "Sub frames" as illustrated in Figure 1: Negative_product_impact, Positive_product_impact, Negative_skin_change, Positive_skin_change and Customer_reaction_to_product. The potential NLSs identified in the previous section such as *shang* 'harm', *sensitive* and *teare* 'chapped hand' are analyzed as target lexical units (LUs) evoking Skin_feel and the corresponding Sub frames. In the customer reviews, the target LUs of `Negative_product_impact` and `Negative_skin_change` are used in negative forms to praise the product (e.g., *hands are not dry*).

[Place Figure 1 near here]

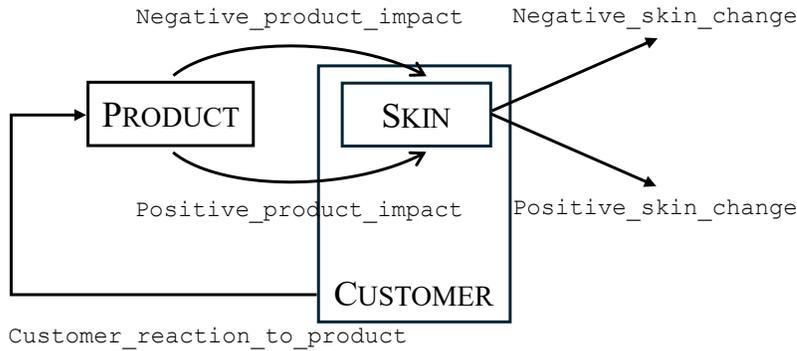

**Figure 1:** The schema for the `Skin_feel` frame.

Different from FrameNet (Ruppenhofer et al. 2016: 15), the transitive and intransitive usages (namely the causative-inchoative alteration: the hand dries vs. dry the hand) are treated as the same frame here. The reason is that semantics rather than syntax should be valued in frame semantics, as discussed above. Causative-inchoative alterations can be seen as different "profilings" of the same event in cognitive grammar (Langacker 1991). Also, change and impact are differentiated in these 5 Sub frames, which have inherent differences. For instance, change-of-state verbs emphasize the result, tending to have causative-inchoative alternations (Alexiadou 2015; Haspelmath 1993). This also supports the treatment of transitive and intransitive usages in the present analysis.

### 3.2.2 Result and discussion

From the same dish soap customer reviews collected in Section 3.1.2, 100 random instances of target LUs evoking `Skin_feel` are identified for each language, through careful manual inspection. The results are summarized in Figure 2. The results are very consistent with those from the automatic BERTopic analysis. Therefore, this fine-grained analysis proves the efficiency and usefulness of the automatic BERTopic, and it verifies the existence and difference of NLSs in crosslinguistic everyday language.

[Place Figure 2 near here]

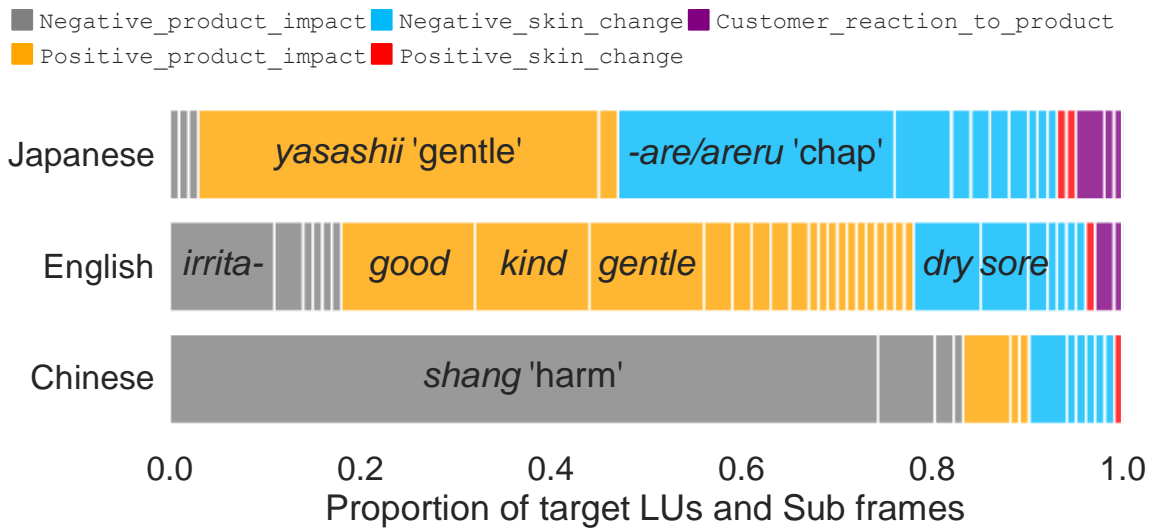

**Figure 2:** Distribution of target LUs and corresponding Sub frames of `Skin_feel` in the dish soap customer reviews. Due to limited space, the proportion of each word is displayed in the figure, but only the names of high-frequency expressions are labeled.

The expressions labeled in Figure 2 all appear in the BERTopic results except *irrita-* and *good*. *Irrita-* has variants of *irritate*, *irritation* and *irritant*, and is thus not counted as a single high-frequency word in BERTopic. Similarly, *good* occurs in multiple topics, so the c-TF-IDF method in BERTopic does not identify it as a distinct topic word for `Skin_feel`. In addition, the frequent word *sensitive* is typically not considered as a target LU here, as it is found in expressions like *good for sensitive skin*, where *sensitive* is not analyzed as impact or change.

Obviously, Chinese favors *shang* 'harm'; English favors *good*, *kind* and *gentle*; Japanese favors *yasashii* 'gentle' and *-are/areru* 'chap'. The results also prove the usefulness of frame semantics for interlingual semantic comparison as discussed in Section 3.2.1. For example, it is actually hard to determine a single translation equivalent for Japanese *yasashii* among words like *gentle*, *mild*, *kind* and *good*. However, frame semantics simplifies interlingual comparison by restricting comparisons to expressions within the same semantic frame, since expressions of different frames are viewed as different expressions. For instance, to see if *-are/areru* 'chap' is more favored in Japanese than its semantic counterparts in Chinese and English, we only need to compare it with expressions belonging to `Negative_skin_change` like *dry* and *sore*. As a result, even if *-are/areru* 'chap' may correspond to *dry*, it is much more frequent than *dry*. And undoubtedly, *shang* 'harm' is more favored in Chinese than its semantic counterparts like *harm* and *kizutsukeru* in English and Japanese. This is because *shang* 'harm' is dominant in `Negative_product_impact` which is of low frequencies in the other two languages.

Again, arbitrary chunks do not seem to be an indisputable reason to explain why different NLSs are favored crosslinguistically. The identified NLSs do not necessarily have fixed patterns and are not necessarily semantically limited to the concept of hand and skin except Japanese *-are*/*areru* 'chap'. For example, *good* can be used in a broad range of semantic fields and in flexible patterns like *good to/for/on my/your/wife's dry/sensitive skin/hands* or simply *We have sensitive skin and this is so good*.

The results also exclude the interference from extralinguistic real-world factors (Bosque 2011; Kjellmer 1991: 114). The results do not support the possibility that an expression is frequent because humans favor that particular frame or that frame occurs more frequently in the real world. In addition, comparing similar concepts in the real world, shows that there are dominant words (which are crosslinguistically different) in each Sub frame.

## 4 Investigating the language-specific prototype of *shang*

In the previous section, an NLS was discovered: Chinese tends to use *shang* 'harm' to describe skin concerns for dish soap. It is important to note that low frequency in a corpus does not necessarily indicate unacceptability (Arppe and Järvikivi 2007). The use of Chinese *shang*'s counterpart *harm* and *kizutsukeru* in this context are acceptable in English and Japanese. Expressions corresponding to *irrita-*, *good*, *kind*, *gentle*, *dry* and *sore* in Chinese (*ciji*, *hao*, *youhao*, *wenhe*, *gan* and *tong*) are acceptable, too. Then, why is *shang* distinctively favored in Chinese? This will be explored in the following sections by revealing its language-specific semantic prototype (in comparison with Chinese synonym *shanghai*, *harm* and *kizutsukeru*). The results aim to validate the onomasiological hypothesis by uncovering the correlation between its prototype and NLS.

### 4.1 Research framework for studying prototypicality

The framework for studying prototypicality is discussed here:

**Definition and treatment of prototype/prototypicality**: Gilquin (2008) points out that there are different yet interconnected definitions for this concept. Among them, the most relevant ones in this study, are the most cognitively salient exemplar among members of a category, and the most frequent item in corpus.

However, the prototypical usage of a word may be under-represented in the available corpus; e.g., *trudge + through … snow* is attested in 30% of the elicited sentences but only 3% of the sentences in the BNC corpus (Dąbrowska 2009). In addition, the prototype is related to features that have high cue validity, applying to a high proportion of members and a low proportion of non-members (Löbner 2013: 274). Thus, relative frequency rather than raw frequency is valued in this corpus study to investigate the prototypes. This will be done by correspondence analysis.

The flexibility of meaning should also be considered (see also the discussion below). In practice, even dictionaries often use fuzzy language like *suggest*, *sometimes* and *often* to describe the words' meanings (Edmonds 1999). Therefore, behavioral profile analysis (Divjak and Gries 2006) will be used to analyze the words' each usage to summarize the overall prototypes.

**Number of prototypes in a word**: The popular view posits only a single prototype in a word, which links the polysemous senses and forms a radiation network (Gilquin 2008). Contrary to it, Fu (2015), Geeraerts (1993, 1997: 61) and Lewandowska-Tomaszczyk (2007) assume that the individual sense of a word can have a prototype. This can be exemplified by Ambridge (2020)'s example: spoons are generally small and metal or large and wooden; but there can not be a prototype that is intermediate in these attributes. In line with the latter approach, this study will identify the senses before studying the prototypes.

**Difficulties in determining senses**: How to determine the senses, especially the "splitting or lumping" or "vagueness vs. polysemy" problem (Geeraerts 1993; Tuggy 1993), is "notoriously problematic" (Gries 2006). Different scholars (Taylor 2012: 223) and dictionaries (Jackson 2002: 88) favor different approaches of these two. Also, every polysemy test is not without problems; see Riemer (2005) for an extensive discussion. Moreover, words' meanings are flexible and absent of clear demarcational boundaries (Geeraerts 2006: 74). This is related to *family resemblance*; see Gisborne (2020: 81) for an example of *climb*. Consequently, the manual annotation of senses is considered "one of the hardest annotation tasks" (Artstein and Poesio 2008).

### 4.2 Determine the usage clusters of *shang* and its synonyms

According to Section 4.1, a word can have multiple prototypes corresponding to its usage clusters. It is necessary to make clear the usage clusters of *shang*, in order to find out which prototype explains its use as an NLS in dish soap customer reviews.

### 4.2.1 Method

Since manual determination of senses is hard and time-consuming, a large-scale, data-driven way is adopted. It also has the merit that "[t]here exists a correct answer that the algorithm must return" (Montes 2021: 1). The underlying principle is "senses as clusters of usages", based on distributionalism which holds that different meanings are associated with different contexts (Lenci 2008). This idea has been widely put into practice by NLP engineers under the term "word sense induction/discrimination" (Nasiruddin 2013; Navigli 2009), where "the number of clusters indicates the number of the target word's senses" (Ghayoomi 2021; see Figure 3 for a visualization example). The same idea can also be seen among linguists like Glynn (2010a, 2016), Gries (2019) and Polguère (2018). The correlation between senses and usage clusters based on count-based embeddings is observed in the works of linguists from QLVL, although they found exceptions due to specific collocational and colligational patterns (De Pascale 2019; Geeraerts et al. 2024; Montes 2021).

In this study, the sense determination is done by clustering prediction-based text embeddings using the multilingual transformer model, LaBSE (Feng et al. 2020). Similar works in the NLP field include Velasco et al. (2022) and Saidi and Jarray (2023). Transformer models have real-world knowledge (Petroni et al. 2019), and approach human-level performance in word sense disambiguation with state-of-the-art results (Coenen et al. 2019; Loureiro et al. 2020).

The multilingual model LaBSE is chosen because it is "language-agnostic". It closely embeds texts of similar meaning regardless of their languages, unlike other multilingual models which may primarily group texts by language (Choenni and Shutova 2020).

**4.2.2 Data collection and processing**

To investigate the nuance of Chinese *shang*'s prototype, comparisons will be made with its Chinese synonym *shanghai* and semantic counterparts in English (*harm*) and Japanese (*kizutsukeru*, whose orthography shares the same kanji as *shang*). To make a fine-grained analysis, only the present negative forms of these 4 words were collected and analyzed because different syntactic forms exhibit different usages (Stubbs 2001; Glynn 2010a, 2010b). The results aim to uncover why the present negative form of Chinese *shang* is more favored to praise the dish soap than other grammatical expressions and its synonyms and counterparts in Chinese, English and Japanese.

The corpus is CC-100 (Conneau et al. 2020) which is based on web-crawled multilingual data of the year 2018 from CommonCrawl (a project that continuously provides a copy of the internet). It is specially filtered to match the style of Wikipedia in order to filter out low-quality web data. It is suitable for crosslinguistic comparison because it uses the same procedures and standards to build the corpora for Chinese, English and Japanese. Also, it is publicly downloadable, allowing customized processing. Web data, especially CommonCrawl, has been widely used for training large language models (Zhao et al. 2023). Notably, web data constitutes 84% of the training data for GPT-3, the former base of the well-known chatbot Chatgpt. To a certain extent, the impressive language skills of these language models, suggest that the web data reflects real-life language usage. In addition, traditional corpora based on authentic materials may exhibit biased usages (Dąbrowska 2009); e.g., customer reviews are not expected from traditional corpora.

The data structure is as follows. From CC-100, 5160 instances of the present negative forms of *shang*, *shanghai*, *harm* and *kizutsukeru* were randomly collected respectively. The text length of each instance was deliberately standardized as follows: Chinese (100 characters), English (60 words) and Japanese (135 characters). This ensures that the texts are roughly equal in length from translation practice.

**4.2.3 Result and discussion**

The collected instances were embedded into semantic space by LaBSE and then reduced to 2 dimensions by UMAP. Finally, they were clustered into 2 clusters by k-means clustering as shown in Figure 3. To understand the criteria underlying the formation of these clusters, the top 5 most frequent objects of these 4 verbs were extracted for each cluster (the percentages indicate the proportion of this usage out of the 5160 total occurrences of the word).

[Place Figure 3 near here]

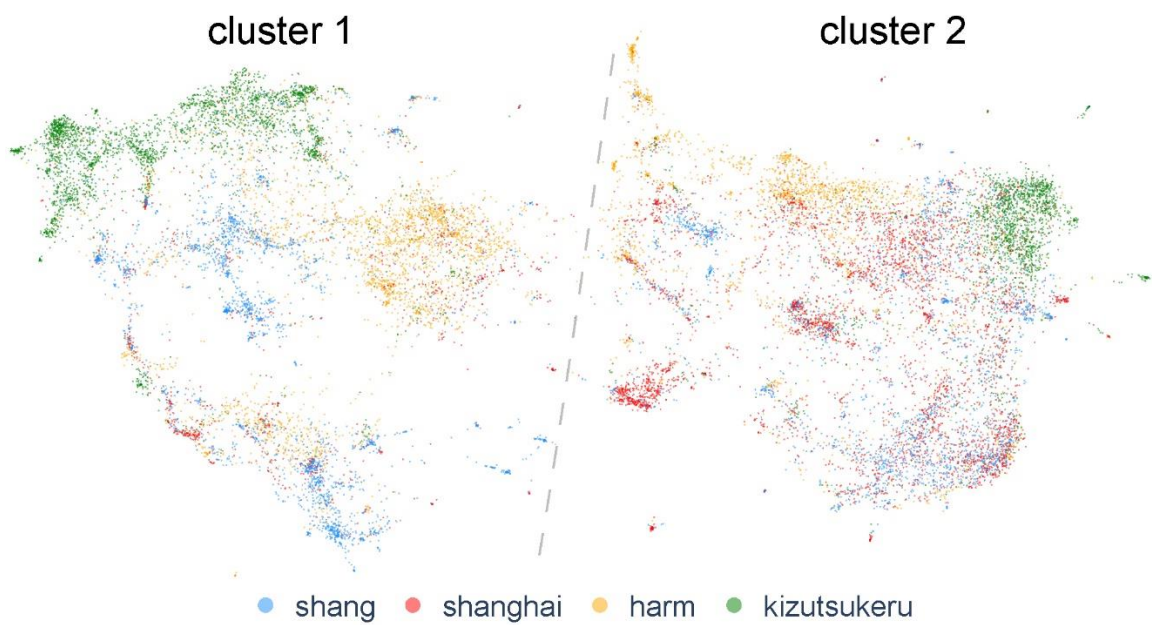

**Figure 3:** 2-D visualization of the usage clusters for *shang*, *harm* and *kizutsukeru*. The two clusters are separated by linear support vector machine.

**Cluster 1**:

Chinese *shang* (51%): *shen* 'body', *shou* 'hand', *shenti* 'body', *yan* 'eye' and *pifu* 'skin'.

Chinese *shanghai* (20%): *shenti* 'body', *renlei* 'human', *pifu* 'skin' and *dongwu* 'animal'.

Japanese *kizutsukeru* (60%): *hada* 'skin', *nikibi* 'pimple', *hifu* 'skin', *toohi* 'scalp' and *yuka* 'floor'.

English *harm* (54%): *environment*, *health*, *skin*, *body* and *animal*.

**Cluster 2**:

Chinese *shang* (49%): *ren* 'person', *ganqing* 'emotion', *xin* 'heart', *shen* 'body' and *zizun* 'self-respect'.

Chinese *shanghai* (80%): *bieren* 'others', *taren* 'others', *ziji* 'self', *liyi* 'interest' and *ren* 'person'.

Japanese *kizutsukeru* (40%): *aite* 'partner', *hito* 'person', *puraido* 'pride', *jisonshin* 'self-esteem' and *jibun* 'self'.

English *harm* (46%): *other*, *anyone*, *computer*, *they* and *you*.

In each cluster, the objects are generally semantically coherent, both intralingually and interlingually except for *computer* in cluster 2 (it is a misclassification due to a local cluster at the periphery of cluster 2). A quick look reveals that cluster 1 is our concern here. It is used as NLS for Chinese in dish soap customer reviews, and it is related to health. Cluster 2 is used for general harm towards people including emotional harm.

Many synonym dictionaries refer to loss/impairment (of any degree) of various aspects such as "value", "function" and "soundness", when describing the synonym group of *harm*, *injure*, *damage* and *hurt* (e.g., Merriam-Webster 1984; Hayakawa 1994); this suggests that the ontological concept of "harm" has a broad semantic range and can be subdivided. Specifically, in this result, cluster 1 is related to functional harm. Cluster 2 is related to emotional harm, destructive harm and interest harm. English examples are shown below. For further examples of Chinese and Japanese, please see Section 4.3.2.

**Cluster 1** (functional harm):

(2) *Natural colors are made only from herbal ingredients and do not <u>harm</u> your health and the environment.*

(3) *Take a clean towel which you can easily use onto your face, and which does not <u>harm</u> your skin texture.*

**Cluster 2**:

(4) *Efforts in watching over our mind, thoughts, actions, and words so that they do not <u>harm</u> ourselves or others.* (emotional harm)

(5) *Most robbers do not <u>harm</u> the victim.* (destructive harm)

(6) *The economic research continues to demonstrate immigrants do not <u>harm</u> the job prospects of U.S. workers.* (interest harm)

Usages of functional harm typically refer to slight or even negligible harm to an affectee's function, which is often not perceptible until the long term. Crabb (1917) describes the synonym group of *harm* as "disadvantage of any person or thing" and *harm* "is the smallest kind of injury"; this reflects that the ontological concept of "harm" can be of slight degree. For illustration, in (2) the environmental harm of hair dye products is indirect, negligible and hardly perceivable; their potential health effects, if any, are likely minor and do not manifest immediately after application. In (3), normally towels are not associated with harm to the skin and wounds, but long-term use may lead to rough skin. In contrast, destructive harm is often associated with a life-threatening meaning and wounds, rather than functional loss.

According to the research framework discussed in Section 4.1, these two clusters correspond to two senses of these words. Although cluster 2 contains three types, this can be attributed to vagueness. For example, nouns about *people* like pronouns are common as the objects of cluster 2, such as *harm someone*, which is vague about the three types of harm and normally not about functional harm.

In the next section, I will closely examine the use of functional harm in cluster 1, which serves as an NLS for Chinese in dish soap customer reviews. The percentage of cluster 1 is around 55% in *shang*, *harm* and *kizutsukeru* while significantly lower in *shanghai*. Therefore, *shanghai* is excluded from further analysis, because the preference for *shang* over *shanghai* in `Negative_product_impact` can be attributed to *shang*'s much higher prototype strength of functional harm. Additionally, this analysis confirms that different syntactic forms exhibit different usages. The above results only display the 5160 usages of Chinese *shang* (in the present negative form) without so-called syntactic "complements" (McDonald 1996), in which cluster 1 accounts for 51%. However, 1011 occurrences of *shang* with complements were also collected, in which cluster 1 accounts for only 25%.

**4.3 Behavioral profile analysis for investigating prototype**

*Shang*, *harm* and *kizutsukeru* are comparable in terms of their prototype strength of functional harm. To uncover the correlation between NLS and *shang*'s prototype, the nuance of *shang*'s prototype of cluster 1 will be examined in detail in comparison with *harm* and *kizutsukeru*.

**4.3.1 Method**

Corpus-driven cognitive semantics can be divided into two methodologies (Glynn 2014). The first one is the mainstream collocation analysis which identifies the co-occurrence of linguistic forms. This study employs the second approach, known as behavioral profile (BP) analysis (Divjak and Gries 2006). BP analysis begins with the manual annotation of each instance of the target expressions using a wide range of features (which are called ID tags). These annotations can then be examined using various statistical techniques (Gries 2010).

Specifically, BP analysis was conducted on 100 randomly selected instances of each verb—*shang*, *harm* and *kizutsukeru* (in their present negative forms), which were collected in cluster 1 in the previous section. In total, 300 instances along with 8700 corresponding ID tags were manually annotated (only 26 instances were misclassified to cluster 1 and thus not chosen for annotation). Finally, a simple correspondence analysis (CA) was performed on the annotation results, which focuses on the relative association (see Section 4.1 for why CA is chosen).

While previous BP studies offer a pre-designed list of ID tags, this study focuses on a series of unique semantic ID tags specifically tailored for this group of verbs of harm. This is because crucial semantic properties may remain unidentified without specially designed ID tags (Proos 2019); meanwhile, the use of highly subjective semantic ID tags has already been justified by Glynn (2010b); finally, it is actually hard for even synonym dictionaries to pre-determine a set of semantic dimensions for differentiating synonyms (Edmonds 1999). Morpho-syntactic ID tags are not the focus of this study, as they may have low discriminatory power in actual synonym analyses (Divjak 2010: 140; Gu 2017; Jarunwaraphan and Mallikamas 2020; Song 2021).

The CA will be visualized using the moon plot which provides clearer visualization than the conventional biplot which is susceptible to misinterpretation (Bock 2011). CA is favored here over multiple correspondence analysis (MCA), which is another frequently used technique in previous BP studies. This is because the focus is on the associations between words and ID tags, which can be recovered from a CA plot with higher accuracy (Greenacre 1991; i.e., only 3 words are involved in this CA, so the explained variance ratio for the first 2 dimensions reaches 100%).

### 4.3.2 Result and discussion

This section will argue that the results shown in Figure 4 indicate the language-specificness of *shang*'s prototype. This finding explains why *shang* is used as an NLS to evaluate the skin concerns of the dish soap. How the linguistic items conceptualize and categorize the real world, does not necessarily align neatly with universal human experience (Hirst 2009). Even synonyms sharing the same denotational range can differ in their prototypes (Geeraerts 1988; Soares da Silva 2015); the group of verbs of harm in this study belongs to this case. The phenomenon of language-specific words has been well analyzed in the literature of Natural Semantic Metalanguage developed by originator Anna Wierzbicka and the colleagues (Goddard 2018; Goddard and Wierzbicka 2014).

Figure 4 shows that the functional harm usages of *shang*, *harm* and *kizutsukeru* are clearly distinguishable. Note that since CA measures relative associations, it is very helpful to check the raw data provided in Appendix. For a concise interpretation of the differences between these terms, *shang* is distinctively associated with everyday self-care for health, *harm* is distinctively associated with social issues like environmental problems, and *kizutsukeru* is distinctively associated with careless/accidental scratches.

[Place Figure 4 near here]

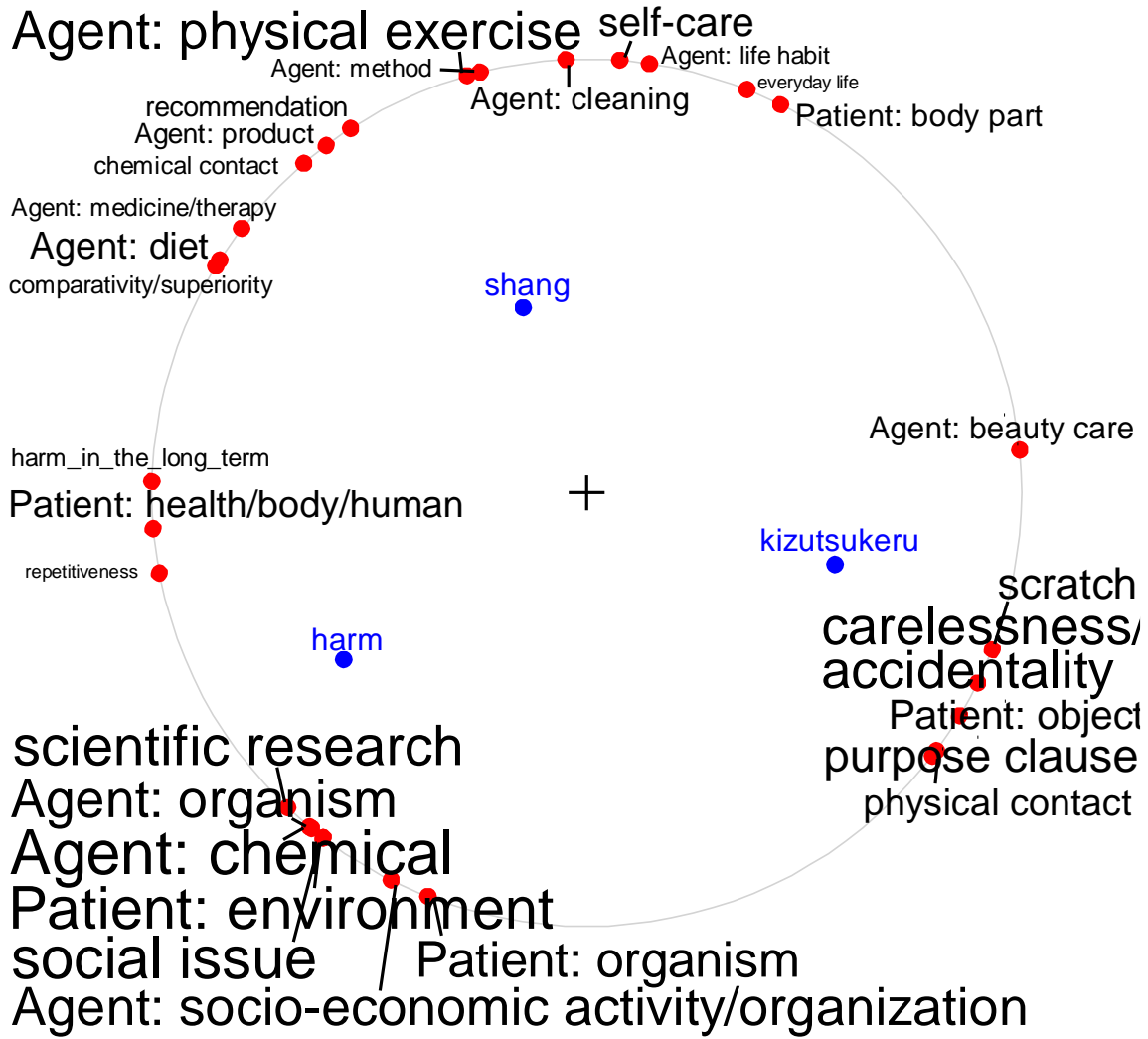

**Figure 4:** Moon plot of the BP analysis results based on CA. The relative associations between words and features are indicated by i) the angle between the origin, the word and the feature (the smaller the stronger), ii) the font size of the feature (the larger the stronger), iii) the distance from the origin to the word (the greater the stronger). For details, please refer to Bock (2011).

(7) *Women de shenghuo zhong you bushao yongcan wuqu, name zenme chi cai neng zai yuezi li rang xin mamamen ji you zugou yingyang gongji baobao naishui er you neng bu <u>shang</u> ziji shenti ne?*

 'There are many misconceptions about eating habits in our lives. So, how can new moms eat during the postpartum confinement to ensure sufficient nutrition for breastfeeding while not <u>harming</u> their own bodies?'

(8) *Tiaosheng bu <u>shang</u> xigai jiu yao jiangjiu fangshi fangfa, zhengque de fangshi, bujin zengjia xiaoguo, hai neng baohu hao xigai.*

'If you want that jumping rope doesn't <u>harm</u> your knees, you need to pay attention to the method. A correct method not only enhances the benefits but also protects your knees.'

(9)  *Hen ganji fumu, women jia san xiongdi congxiao de ganmao dou chi zhongyao, suiran man, danshi daozhi women san xiongdi changwei gongneng dou tebie hao. Ta zhishao shi bu <u>shang</u> gan bu <u>shang</u> wei de, shi zhide tuiguang de.*

'I am very grateful to our parents. In our family, the three brothers have been taking traditional Chinese medicine for colds since childhood. Although it's slow, it has resulted in all three of us having particularly good gastrointestinal function. At the very least, it doesn't <u>harm</u> the liver or stomach, and it's worth promoting.'

(10) *It is a safe replacement for petroleum, as it is non-toxic and doesn't <u>harm</u> the environment.*

(11) *The eggs of the fruit fly are not poisonous and do not <u>harm</u> us.*

(12) *Kikinzoku ya megane o fuku sai ni shiyoo sareru koto ga ooku, kuruma o <u>kizutsuke</u>-nai koto de ninki desu ga, kono seemu gawa ni wa dono yoo na meritto/demeritto ga aru no deshoo ka.*

'It is often used for wiping precious metals and glasses, and it is popular because it doesn't <u>harm</u> cars, but what are the advantages and disadvantages of this chamois leather?'

(13) *Tadashi rediisu sheebaa ya kamisori o tsukau toki wa hada o <u>kizutsuke</u>-nai yoo ni chuui shinagara okonatte kudasai.*

'However, when using a lady shaver or razor, please proceed carefully to not <u>harm</u> your skin.'

Below is a detailed interpretation and discussion of the results. Only the functional harm usages in the present negative forms are analyzed, so the overt mention of this condition is omitted. However, in fact these results can still be applied to the affirmative form of *shang* without complements.

The core of Chinese *shang*'s prototypical meaning is closely linked to everyday self-care for health. It is associated with the ID tag "everyday life" and related agents like "life habit", "diet", "cleaning", "physical exercise" and "product"; additionally, "repetitiveness" and "harm_in_the_long_term" are related to everyday life. Furthermore, it is associated with the ID tag "self-care", which is about ways to take care of health. This is related to ID tags like "Agent: method", "comparativity/superiority" and "recommendation", which are about the ways. Health-related ID tags include "Agent: medicine/therapy" and "Patient: body part". Actually, the ID tags about everyday life and self-care are interrelated, because self-care is about what we choose to do in everyday life routines.

(7) and (8) are examples of Chinese *shang*. They recommend everyday self-care tips that teach the correct life habits for eating and exercising. When eating and exercising, we have to choose the foods and methods. However, we are concerned that if we do not follow the right practices, these routines could harm our health in the long term. Normal ways of breastfeeding and jumping rope may be harmful to our health and knees to a degree, especially in the long term. Therefore, by using *shang*, better ways which avoid or minimize the harm are recommended instead.

In summary, when we say *A bu shang B* 'A does not <u>harm</u> B', we typically mean the following:

a. we choose how to do things and what to use in everyday life habits (e.g., what to eat and what products to use).

b. B is usually a body part.

c. poor choices result in worsening the function/health of B in the long term.

d. we want to maintain good body function/health.

e. A is recommended because it does not harm B's function/health and is considered an appropriate and better choice.

In addition, there is an intriguing aspect of the usage of *shang*, further supporting its culture-specific nuances. It is commonly used to describe the "harm" in the ideologies and concepts of traditional Chinese medicine, even in everyday conversations like (9). For example, in this study, the ten most frequent objects of *shang* include *wei* 'stomach' and *gan* 'liver', which are very rarely seen in the objects of *shang*'s Chinese synonym *shanghai*, not to mention English word *harm* and Japaness *kizutsukeru*. *Wei* 'stomach' and *gan* 'liver' belong to *zangfu* in the theory of traditional Chinese medicine. Notably, these *zangfu* do not correspond to their anatomical counterparts in modern western medicine. Instead, they are primarily viewed as certain "functions" of the body (Liao et al. 2017). Traditional Chinese medicine is characterized by its philosophy like yin and yang, and treatments like food therapy and herbal medicine (Hsu 2018). Health cultivation is a key concept of it, which encourages ordinary people to cultivate health and harmony through daily activities. This corresponds to the meaning of everyday self-care. Dear (2012) has illustrated the situation of *yangsheng* 'health cultivation' with this example: "an unlucky or foolish dietary choice can lead to acute pain and discomfort. Casual or careless abuse over a lifetime will eventually lead to chronic illness in the whole body system". This illustration aligns closely with the interpretation of the scenarios of *shang* in the present analysis. In summary, this paper's interpretation of *shang*'s nuance and its language/culture-specificness are supported by *shang*'s relevance with traditional Chinese medicine.

The usage of Chinese *shang* can be compared with *harm* and Japanese *kizutsukeru*. *Harm* is found in scientific discourses and discussions of social issues like (10), which are rarely seen in *shang* and *kizutsukeru*. ID tags related to the scientific discourse are "scientific research", "Agent: organism", "Patient: organism" and "Agent: chemical". ID tags related to discussions of social issues include "social issue", "Agent: socio-economic activity/organization" and "Patient: environment". However, this does not indicate that the prototypical meaning of *harm* is strictly limited to these domains or a formal register, since the raw percentages of these ID tags are actually not overwhelmingly high as shown in Appendix. Instead, it suggests that the meaning of *harm* is more abstract and general, not carrying strong pragmatic nuances like *shang*. For example, while (11) is about harm towards human health, it is not about everyday self-care.

Japanese *kizutsukeru* is typically associated with careless/accidental scratches. It can be used for the surface of objects like (12) and also the human body like (13) which is often about harm to skin during beauty care. However, in this latter case, unlike *shang*, its harm to the skin focuses on scratches caused by physical contact with hard objects. *Kizutsukeru* frequently appears in "purpose clause" together with terms like *chuui* 'pay attention'. This is also related to its property of "carelessness/accidentality". Its presence in a purpose clause implies that the harm and the action described by *kizutsukeru* can be controlled and prevented through careful attention. In this case, the pragmatic effect of *kizutsukeru* is about a cautionary reminder rather than "recommendation" or "comparativity/superiority".

While Bp analysis relies on manual and subjective annotations on a sampled corpus, the results from this Bp analysis can be supported by automatic extraction of the objects and keywords—using Egbert and Biber (2019)'s approach. Some noteworthy words of them are referred to below. *Shang*'s topic words include *jianfei* 'lose weight', *he* 'drink', *yundong* 'physical exercise' and *chi* 'eat', which are important activities that require a great deal of attention in everyday self-care. In contrast, English topic words contain *environment*, *chemical*, *plant*, *animal*, *eco*, *energy*, *organic*, *sustainable* and *cell*; these words and the themes behind them align with the BP analysis results and are rarely found in *shang*'s usages. *Kizutsukeru*'s topic words contain *nikibi* 'pimple', *kosuru* 'rub', *yoo* 'in order to', *chuui* 'pay atterntion', *kudasaru* 'please' and *datsumoo* 'depilation'; and its frequent objects include *nikibi* 'pimple', *yuka* 'floor', *nenmaku* 'mucosa', *kabe* 'wall' and *hyoomen* 'surface'. Obviously, these words are about cautionary reminders and activities involving careless/accidental scratches on surfaces.

By comparing interlingually with words denoting the concept of "harm", the analysis reveals the language-specific prototype of Chinese *shang* due to its special pragmatic nuances. It can be argued that because *shang*'s prototypical meaning is related to everyday self-care, it is used as a nativelike selection for commenting on the skin feel of dish soap products by Chinese native speakers. This case study then supports the correlation between nativelike selection and prototypicality.

Specifically, the dish soap is a household necessity used in daily routines. The skin feel is not the core concern when customers choose dish soaps because their harm is negligible. However, we use them on a daily basis, and are worried that they may result in worsening the function/health of the skin in the long term (e.g., roughness, dryness, cracking, redness, sensitivity and unattractive appearance). Therefore, customers may seek better dish soap products that minimize such harm. In summary, choosing the right dish soap, namely products that *bu shang shou* 'does not harm hands', is about everyday self-care, which explains why *shang* is favored here.

## 4.4 Testing the predictive power of the analysis

When establishing the semantic motivation behind NLSs, rather than dismissing them as arbitrary chunks, it is important to worry about the existence of counterexamples (Alonso Ramos 2017). To test whether the preference for Chinese *shang* can be predicted under the onomasiological hypothesis to be generalized to other discourses, customer reviews of e-readers (e.g., Amazon Kindle) and negative ion hair dryers were collected and analyzed. Their selling point is their concern for the customers' eyes or hair. These products are related to *shang*'s prototypical meaning of everyday self-care, because people use electronic screens and hair dryers in daily life, and have health concerns for their eyes and hair. A simple analysis was performed on the collected data by counting the frequencies of the target words (i.e., words referring to eyes or hair) and the frequencies of *shang*, *harm* and *kizutsukeru* which are used to comment on the products' impact on eyes or hair, as shown in Table 1.

[Place Table 1 near here]

**Table 1:** Analysis of *shang*'s favoredness in additional customer reviews by counting the raw frequencies.

|  | E-reader | | Negative ion hair dryer | |
|---|---|---|---|---|
| Chinese | *yan/yanjing* 'eye' | *shang* | *toufa/fa* 'hair' | *shang* |
|  | 425 | 73 | 526 | 51 |
| English | *eye* | *harm* | *hair* | *harm* |
|  | 105 | 0 | 789 | 0 |
| Japanese | *me* 'eye' | *kizutsukeru* | *kami* 'hair' | *kizutsukeru* |
|  | 189 | 0 | 520 | 0 |

(14) *It has the convenience of digital copies with an extremely long-lasting battery and a screen that does not <u>harm</u> or tire your eyes. I truly recommend it.* (reddit.com)

(15) *I bought this Hair dryer for my daughter. Sometimes I use it myself. Quieter and lighter than conventional hair dryers. It does not <u>harm</u> the hair at all. It dries very easily.* (amazon.com)

(16) *Denshi shoseki riidaa yori mo keitai denwa o shiyoo shite iru node hanbai shitai node, me o <u>kizutsuke</u>-zu ni yoru no manga o yomu no ni teki shita bakkuraito o sonaete imasu.* (from Google search)

 'Since I am using my mobile phone more than this e-book reader, I want to sell this device equipped with a backlight suitable for reading manga at night, which does not <u>harm</u> the eyes.'

(17) *Tada kawakasu dake de wa nai, shinhassoo no doraiyaa. Teiondo, teifuuryoo de kami no ke o <u>kizutsuke</u>-zu, shikkari to kawakashimasu.* (from Google search)

'Not just drying the hair, but a dryer based on a new idea. It dries thoroughly but does not <u>harm</u> your hair with low temperature and low airflow.'

The results are consistent with the prediction of the onomasiological hypothesis, which suggests the correlation between nativelike selection and prototypicality. This analysis at least shows that *shang* remains preferred over its semantic counterparts in English and Japanese, in texts beyond dish soap customer reviews. The use of *shang* is common in these additional materials, whereas *harm* and *kizutsukeru* are infrequent and even not attested in the collected data. Note that their non-attestation does not mean unacceptability; examples of them are found on the internet such as (14)-(17).

As discussed above, the results support the semantic motivation behind the favoredness of *shang* rather than arbitrariness and chunkedness. *Shang* frequently combines with not only *shou* 'hand' but also with *yan/yanjing* 'eye' and *toufa/fa* 'hair', with each pair associated with different sources of harm (i.e., chemical, light and heat); what they have in common is the theme of everyday self-care. Finally, it is interesting that while the use of *harm* and *kizutsukeru* is acceptable, they did not appear in the collected data. This suggests that the lexical choices made by native speakers are not random choices among acceptable expressions, implying the existence of motivations and regularity behind them.

However, the frequency of *shang* might not seem overwhelmingly high; this might be attributed to additional factors such as contextual and pragmatic factors (Geeraerts et al. 1994; Grondelaers and Geeraerts 2003) that also influence NLS. For instance, Chinese *shunhua* 'smooth' appeared 79 times in the collected customer reviews of negative ion hair dryers to describe their impact on customers' hair; *shunhua* 'smooth' might be favored apart from *shang* because smoothness of the hair is overtly visible and directly related to the effects of negative ions, thus capturing more attention from speakers. In addition, there might exist the phenomenon of *competition* (MacWhinney et al. 2014; Masini 2019) in the lexical choices; for example, *hu yan* 'protect eyes', which appeared 90 times in the collected customer reviews of e-readers, might be a collocation and compete with the use of *shang*. Nonetheless, the interlingual comparison results support the correlation (not determination) between prototypicality and NLS, while leaving room for discussions of other factors.

## 5 Conclusion: Non-collocational nativelike selection

Section 3.1.3 reveals that, apart from collocations, non-collocational nativelike selections are not rare in everyday language. There are two reverse directions in the study of lexical choices: "from argument/base to predicate/collocate" and "predicate-driven" (Almela-Sánchez 2019). If analyzing from argument to predicate, the analysis will begin with the argument word, for example, *shou* 'hand', and find that the predicate word *shang* 'harm' is preferred to combine with it, over other possible predicate words such as *hao* 'good' and *gan* 'dry'. If the analysis stops here, it might be concluded that this nativelike selection (i.e., *shang shou* 'harm the hand') is an arbitrary collocation.

However, this study contributes by also examining the reverse direction, from predicate to argument, analyzing *shang*'s entire usages beyond its combination with *shou* 'hand', in order to find the semantic motivation behind this nativelike selection. By revealing the language-specific prototype of *shang*, this study provides evidence for the correlation between nativelike selection and prototypicality, specifically the onomasiological hypothesis: "A referent is more readily named by a lexical item if it is a salient member of the category denoted by that item" (Grondelaers and Geeraerts 2003: 74).

This study aims to provide a preliminary analysis of nativelike selection from an onomasiological perspective through interlingual comparisons, which was seldom explored in previous research despite its significance in language learning and research. To facilitate this aim, innovative methods such as the use of semantic embeddings are designed to discover nativelike selections and explore prototypes, which can be used in future studies. Regarding future research, Bosque (2011) states that most collocations are not arbitrary; however, whether most NLSs are arbitrary or can be analyzed through the correlation with prototypicality is beyond the scope of this study. Furthermore, the robustness of this onomasiological hypothesis is supported in Section 4.4, but it also suggests the existence of other factors influencing nativelike selection. These can be explored in future research.

# Appendix: Input data for the correspondence analysis

| Type of ID tag | ID tag | Target word | | |
|---|---|---|---|---|
| | | Shang | Harm | Kizutsukeru |
| Agent | life habit | 0.76 | 0.3 | 0.54 |
| | method | 0.44 | 0.19 | 0.24 |
| | product | 0.42 | 0.22 | 0.17 |
| | diet | 0.2 | 0.13 | 0.01 |
| | beauty care | 0.2 | 0.05 | 0.35 |
| | cleaning | 0.14 | 0.04 | 0.07 |
| | medicine/therapy | 0.14 | 0.09 | 0.07 |
| | physical exercise | 0.11 | 0.01 | 0 |
| | socio-economic activity/organization | 0 | 0.23 | 0.07 |
| | chemical | 0 | 0.12 | 0 |
| | organism | 0.01 | 0.08 | 0.01 |
| Patient | body part | 0.63 | 0.11 | 0.49 |
| | health/body/human | 0.27 | 0.31 | 0.03 |
| | environment | 0 | 0.36 | 0.02 |
| | organism | 0 | 0.21 | 0.09 |
| | object | 0.09 | 0.08 | 0.41 |
| Syntactic position | purpose clause | 0.03 | 0.1 | 0.68 |
| Semantic feature | physical contact | 0.19 | 0.23 | 0.91 |
| | scratch | 0.15 | 0.05 | 0.86 |
| | carelessness/accidentality | 0.01 | 0 | 0.59 |
| | chemical contact | 0.25 | 0.14 | 0.11 |
| | repetitiveness | 0.91 | 0.79 | 0.64 |
| | harm_in_the_long_term | 0.79 | 0.7 | 0.38 |
| Theme | everyday life | 0.92 | 0.42 | 0.77 |
| | self-care | 0.8 | 0.09 | 0.35 |
| | recommendation | 0.81 | 0.41 | 0.35 |
| | comparativity/superiority | 0.84 | 0.56 | 0.37 |
| | social issue | 0 | 0.18 | 0.01 |
| | scientific research | 0.02 | 0.21 | 0 |

wards automatic construction of Filipino WordNet: Word sense induction and synset induction using sentence embeddings, *arXiv*, *Preprint arXiv:2204.03251*.

Walker, Crayton. 2008. Factors which influence the process of collocation. In Frank Boers & Seth Lindstromberg (eds.), *Cognitive linguistic approaches to teaching vocabulary and phraseology*, 291–308. Berlin: De Gruyter Mouton.

Wray, Alison. 2002. *Formulaic language and the lexicon*. Cambridge: Cambridge University Press.

Wray, Alison. 2012. What do we (think we) know about formulaic language? An evaluation of the current state of play. *Annual Review of Applied Linguistics* 32. 231–254.

Yong, Zheng Xin, Patrick D. Watson, Tiago Timponi Torrent, Oliver Czulo & Collin Baker. 2022. Frame shift prediction. In Nicoletta Calzolari, Frédéric Béchet, Philippe Blache, Khalid Choukri, Christopher Cieri, Thierry Declerck, Sara Goggi, et al. (eds.), *Proceedings of the thirteenth language resources and evaluation conference*, 976–986. Marseille, France: European Language Resources Association.

Zaytseva, Victoria. 2016. *Vocabulary acquisition in study abroad and formal instruction: An investigation on oral and written lexical development*. Barcelona, Spain: Universitat Pompeu Fabra dissertation.

Zhao, Wayne Xin, Kun Zhou, Junyi Li, Tianyi Tang, Xiaolei Wang, Yupeng Hou, Yingqian Min, et al. 2023. A survey of large language models, *arXiv*, *Preprint arXiv:2303.18223*.